\documentclass[runningheads]{llncs}

\usepackage[year=2024,ID=1014]{accv}
\usepackage{accvabbrv}

\usepackage{graphicx}
\usepackage{booktabs}

\usepackage{animate}
\usepackage{subcaption}
\usepackage{multirow}
\usepackage{array}
\usepackage{cite}
\usepackage{amsfonts}
\usepackage{float}
\usepackage{verbatim}
\usepackage{amsmath}
\DeclareMathOperator*{\argmin}{arg\,min}
\usepackage[accsupp]{axessibility}  % Improves PDF readability for those with disabilities.

\usepackage[pagebackref,breaklinks,colorlinks,citecolor=accvblue]{hyperref}
\usepackage{orcidlink}

\begin{document}

% ---------------------------------------------------------------
% TODO REVIEW: Replace with your title
\title{Enhancement of 3D Gaussian Splatting using Raw Mesh for Photorealistic Recreation of Architectures} 

% TODO REVIEW: If the paper title is too long for the running head, you can set
% an abbreviated paper title here. If not, comment out.
\titlerunning{Abbreviated paper title}

% TODO FINAL: Replace with your author list. 
% Include the authors' OCRID for the camera-ready version, if at all possible.
\author{Ruizhe Wang \and
 Chunliang Hua \and
 Tomakayev Shingys \and
 Mengyuan Niu \and
 Qingxin Yang \and
 Lizhong Gao \and
 Yi Zheng \and
 Junyan Yang \and
 Qiao Wang\orcidlink{0000--0002-5271-0472}}

% TODO FINAL: Replace with an abbreviated list of authors.
\authorrunning{Wang et al.}
% First names are abbreviated in the running head.
% If there are more than two authors, 'et al.' is used.

% TODO FINAL: Replace with your institution list.
\institute{}

\footnotetext[1]{R. Wang, C. Hua, T. Shingys,  Q. Yang and L. Gao was with the School of Information Science and Engineering, Southeast University, Nanjing, Jiangsu, China(email: rz\_wang@seu.edu.cn, chunlianghua@seu.edu.cn, eintotonello@gmail.com, myniu@seu.edu.cn, lzgao@seu.edu.cn)}

\footnotetext[2]{Q. Yang, Y. Zheng and J. Yang was with the School of Architecture, Southeast University, Nanjing, Jiangsu, China(email: qingxinyang@seu.edu.cn, yizheng@seu.edu.cn, yangjy\_seu@163.com)}

\footnotetext[3]{Q. Wang was with both School of Information Science and Engineering and the School of Economics and Management,  Southeast University, Nanjing, Jiangsu, China(Corresponding author, email:qiaowang@seu.edu.cn)}

\maketitle

\begin{abstract}
  The photorealistic reconstruction and rendering of architectural scenes have extensive applications in industries such as film, games, and transportation. It also plays an important role in urban planning, architectural design, and the city's promotion, especially in protecting historical and cultural relics. Due to better performance than NeRF, 3D Gaussian Splatting has become a mainstream technology in 3D reconstruction. Its only input is a set of images but it relies heavily on geometric parameters computed by the SfM process. At the same time, there is an existing abundance of raw 3D models, that could inform the structural perception of certain buildings but cannot be applied. In this paper, we propose a straightforward method to harness these raw 3D models to guide 3D Gaussians in capturing the basic shape of the building and improve the visual quality of textures and details when photos are captured non-systematically. This exploration opens up new possibilities for improving the effectiveness of 3D reconstruction techniques in the field of architectural design.
\end{abstract}

\section{Introduction}
\label{sec:intro}

3D reconstruction of architectural scenes, which combines photography and image processing methods, enables the creation of accurate, photo-realistic models of buildings \cite{8446956, frahm20093d}. This technique has wide-ranging applications, from aiding architectural design and urban data analysis to preserving cultural heritage \cite{bent2022practical}, creating virtual spaces for AR and VR tourism, and contributing to digital gaming and filming industries.

Recent advancements in computer vision and machine learning have led to innovative virtual imaging techniques such as the Neural Radiance Field (NeRF) \cite{mildenhall2020nerf} and its subsequent developments like Mip-NeRF \cite{Barron_2021_ICCV}, Mip-NeRF360 \cite{Barron_2022_CVPR}, GRF \cite{Trevithick_2021_ICCV}, and PixelNeRF \cite{Yu_2021_CVPR}. These techniques have resolved issues like aliasing, introduced 360-degree scenes, and implemented CNN encoding for image features. Further improvements by NeRFBusters \cite{Warburg_2023_ICCV} and DreamFusion \cite{poole2022dreamfusion}, are achieved by adding data-driven priors. InstantNGP \cite{10.1145/3528223.3530127} and TensoRF \cite{Chen2022ECCV} have improved training speed, memory optimization, and accuracy by replacing MLP with other 3D representations. The latest method, 3D Gaussian Splatting (3DGS) \cite{kerbl3Dgaussians}, achieves real-time rendering and represents a scene as a set of millions of three-dimensional Gaussians. SuGaR \cite{guedon2023sugar} then proposed a method of extracting surface mesh from 3DGS, providing great inspiration for further research.

Nonetheless, challenges may emerge when employing 3DGS for rendering larger-scale buildings and scenes \cite{chen2024survey}. The quality of photographic data collected by drones around the object can be affected by various limitations such as drone battery life, camera shooting angle constraints, or unforeseen obstacles within the scenes. This can result in artifacts in the generated point cloud, manifesting as holes, floaters, and blurry surfaces. This problem, similar to the large viewpoint change between the training sequence and evaluation sequence discussed in NeRFBusters, shows that the performance of NeRF-like methods, including 3DGS, becomes fragile in such instances.

\begin{figure}
    \centering
    \includegraphics[width=0.8\linewidth]{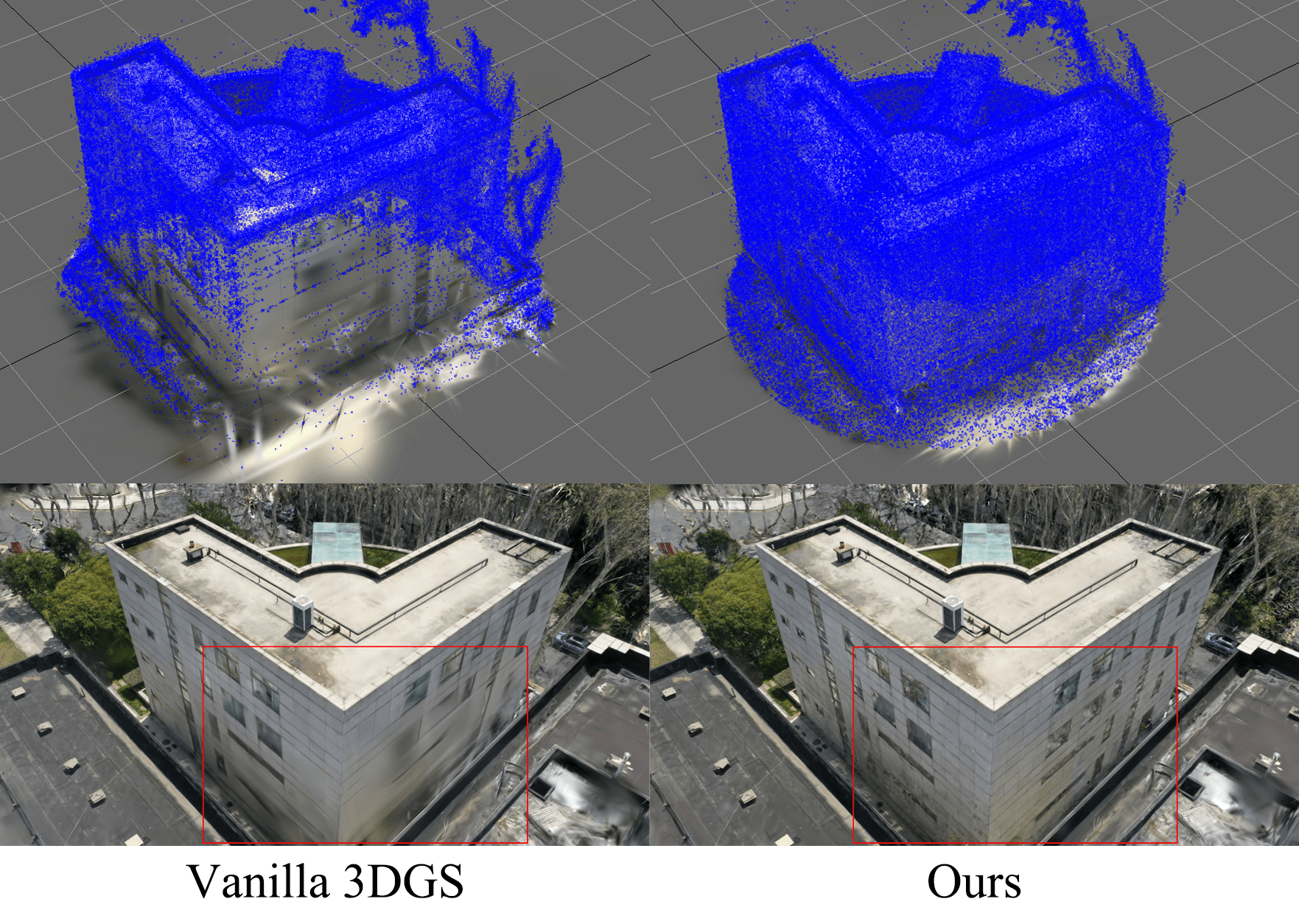}
    \caption{A comparison between vanilla 3DGS and our raw model enhanced method. The centers of the Gaussians after training are demonstrated in the first row. Vanilla 3DGS fails to represent the basic geometry and texture of the wall due to inadequate observation, but ours succeeds. The rendered scene is shown in the second row. Please note the windows in the red box, which denote our enhanced performance in capturing realistic details.}
    \label{fig:fig1}
\end{figure}

\begin{figure}
    \centering
    \includegraphics[width=0.7\linewidth]{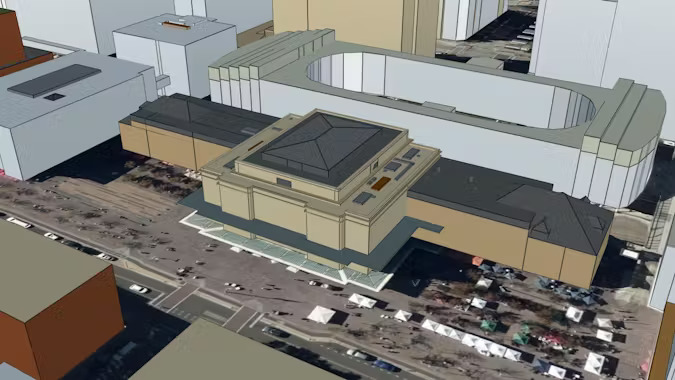}
    \caption{Example of raw 3D models from Cesium OSM Buildings (figure from Cesium website). The raw models already demonstrate the basic shape of the buildings, but obviously without detailed appearance.}
    \label{fig:fig2}
\end{figure}

Our research aims to address this issue for architectural scenes by introducing prior shape data, specifically coarse mesh. Mesh is a common data type used in object modeling, where the surface triangles represent the shape of an object. It can also contain the color features of an object but is typically a raw model for buildings. A simple, low-precision structural mesh encompassing all mapped buildings worldwide, can be exported from Geographic Information System (GIS) data. A slightly more precise mesh of the majority of famous buildings can be obtained from open 3D global content websites like Cesium\footnote{https://cesium.com/platform/cesium-ion/content/}, shown in \cref{fig:fig2}. 

We sample point clouds on their surface and align them with the sparse point cloud calculated by COLMAP, which is the initial position of Gaussians, and use the merged point cloud as prior shape information. This process frees 3DGS from fitting the basic geometry of the building and these priors can enhance the learning quality of 3DGS for generating detailed geometric shapes and surface textures. Our proposed method has been applied to four scenes, resulting in notable enhancements in both the shape and texture quality. This highlights the potential of our approach to significantly improve the visual quality of 3DGS renderings within the architectural industry.

To summarize, we make contributions as follows:
\begin{itemize}
    \item We introduce a pipeline that incorporates existing 3D models, specifically coarse meshes, to guide the initial point cloud sampling. This serves as both a supplement and correction to the geometric parameters calculated by SfM.
    \item We propose a procedure to allocate varying numbers of points to triangles based on their area and a method of assigning initial color, ensuring a more accurate representation of the mesh surface.
    \item We apply this integration in four scenes with photos captured in a non-systematic manner and demonstrate its superiority over vanilla 3DGS.
\end{itemize}

In the remainder of the paper, we discuss related work, give a brief overview of recent advancements in radiance fields, describe our approach, and compare it to the vanilla 3D Gaussian Splatting.

\section{Related Work}
Traditional methods of 3D reconstruction such as laser scanning and photogrammetry, including Structure-from-Motion (SfM) \cite{Schonberger_2016_CVPR} and multi-view stereo (MVS) \cite{4408933}, have been complemented by novel methods based on radiance fields and view synthesis. The Neural Radiance Field (NeRF) is a significant development in this area, offering detailed reconstruction and novel view composition. Follow-up works like Block-NeRF \cite{Tancik_2022_CVPR} have extended NeRF’s capabilities to reconstruct large scenes, albeit requiring substantial image datasets. However, NeRF has limitations concerning datasets, training and rendering speed, result editability, and dataset size requirements.

3D Gaussian Splatting (3DGS) is one of the most advanced rendering techniques based on radiance fields to overcome the limits of NeRF. Thanks to the explicit representation of 3D scenes, it has state-of-the-art performance in accuracy and fast training within a few minutes, achieving real-time rendering for the first time in this field. This efficient representation is a set of millions of three-dimensional Gaussians, which resemble small ellipsoids with different centers, scales, rotations, opacity, and colors when viewed from different angles. The Gaussians are defined by a full 3D covariance matrix \(\mathbf{\Sigma} \in \mathbb{R}^{3 \times 3} \) in world space and centered at point (mean) \(\mu \in \mathbb{R}^3 \), spherical harmonic (SH) coefficients, and an opacity value \(\alpha \in [0, 1)\), as follows:
\begin{equation}
g(x) = e^{-\frac{1}{2}(x - \mu)^T\mathbf{\Sigma}^{-1}(x - \mu)}.
\end{equation}
Given a scaling matrix \( \mathbf{S} \), rotation matrix \( \mathbf{R} \), and viewing transformation \( \mathbf{W} \), they define the corresponding covariance \( \mathbf{\Sigma} \) and projected covariance \( \mathbf{\Sigma}' \):
\begin{equation}
\mathbf{\Sigma} = \mathbf{RSS}^T\mathbf{R}^T,
\end{equation}
\begin{equation}
\mathbf{\Sigma}' = \mathbf{JW \Sigma W}^T\mathbf{J}^T.
\end{equation}
where \( \mathbf{J} \) is the Jacobian of the affine approximation of the projective transformation.

The scene is rendered through splatting rasterization, which is accomplished by transforming the 3D Gaussians into 2D Gaussians on the image plane, simply by skipping the third line and column of \(\mathbf{\Sigma}'\). Then, to render the color \( C(p) \) of pixel \( p \), 3D Gaussian Splatting executes \(\alpha\)-blending, multiplying the Gaussians with corresponding \(\alpha\) according to its depth. For instance, \( C(p) \) can be calculated as follows after sorting:
\begin{equation}
    C(p) = \sum_{i=1}^{N} c_i \alpha_i G'_i(p) \prod_{j=1}^{i-1} (1-\alpha_j G'_j(p)),
\end{equation}
where \( c_i \) is the color value of each anisotropic Gaussian from a certain direction calculated from SH coefficients, and \( G'_i \) is the 2D Gaussian after projection. This leads to a highly efficient rendering process. Then the algorithm compares the rendered image to the ground truth and calculates the loss, which is applied to optimize the virtual 3D scene. The example presented by SuGaR proposed a method of extracting surface mesh from 3D Gaussians, re-binding 3D Gaussians with mesh patches for refinement, and finally using the former to render the mesh model, technically utilizing rendering to optimize the original mesh, giving us great inspiration for further research.

\section{Methodology}
Our work draws inspiration from the joint refinement step of the SuGaR implementation, which utilizes a coarse mesh as a geometric prior and generates Gaussians on its surface. However, the joint refinement of the SuGaR model cannot be directly applied to the raw model of buildings due to two reasons:
\begin{enumerate}
    \item The quantity and shape of the building's coarse mesh surface are rudimentary and do not meet the mesh precision required for SuGaR refinement. Furthermore, the SuGaR model itself cannot split the mesh surface during learning;
    \item The SuGaR method sets a fixed number of Gaussians for each surface, typically one or six, which cannot accommodate the vast area differences of the building's mesh surfaces.
\end{enumerate}
These factors lead to the inability of the SuGaR method's refinement to correctly learn the structure and color of the building when importing prior shape information. We revert to the original 3DGS and import the building's raw mesh model to find a solution.

The main component of our method is the initial point cloud sampling. This step is inspired by the process of binding Gaussians to the triangles of the mesh in the SuGaR model. We define the barycentric coordinates \( \mathbf{w} = (w_a, w_b, w_c )^T \in [0, 1]^3 \) similarly to represent the position of the sampled points. The advantage of the barycentric coordinates is that there is no need to consider the real coordinates of the triangle vertices in space, but only a one-hot vector in advance. This allows our program to do parallel point sampling for triangles, making it extremely fast. However, unlike SuGaR, which assigns the same number of Gaussians to each triangle, we assign different numbers of points to surfaces of each size.

To elaborate, we first sort the triangles according to their area, then grade them by area size from largest and going in descending order, with a grading area ratio threshold of 1/4. Then, commencing with the triangle of the smallest area, we allocate the number of sampled points for each grade, following an exponential increment of 4. We limit the number of grades to \(N + 1 = 6, 7, 8, 9\) for different scenes to generate a reasonable number of sampled points. The assignment of levels of the triangles and the corresponding number of sampled points are listed in \cref{tab:Table1}. The positions of these sampled points are determined as the geometric centers of the diminutive triangles, incessantly subdivided into 4 by inserting a new vertex at the midpoints of each side of the triangle. A recursive procedure outputs the barycentric coordinates of those centers, which are stored for direct recall in subsequent operations. The final positions of sampled points for each grade are calculated as follows:
\begin{equation}
    \mathbf{P}_{nm} = \mathbf{W}_n\mathbf{V}_m^T, 
\end{equation}
where \( \mathbf{W}_n = (\mathbf{w}_1, \mathbf{w}_2, ..., \mathbf{w}_{4^n})^T, n = 0, 1, 2, ..., N \) is an \(4^n \times 3\) matrix of barycentric coordinates of each grade, and \( \mathbf{V}_m = (\mathbf{v}_{ma}, \mathbf{v}_{mb}, \mathbf{v}_{mc})\) is a \(3 \times 3\) matrix of real positions of vertex A, B, C of the \(m\)th triangle. Each row of the output \(\mathbf{P}_{nm}\) represents the 3D coordinates of a sampled point.

\begin{table}[h]
\caption{The grade and number of sampled points of triangles (when \(N + 1 = 9\))}
\label{tab:Table1}
\centering
{
\renewcommand{\arraystretch}{1.5}
\begin{tabular}{cm{1.2cm}<{\centering}m{1.2cm}<{\centering}m{1.2cm}<{\centering}m{1.2cm}<{\centering}m{1.2cm}<{\centering}}
\toprule
Grade & 0 & 1 & ... & 7 & 8 \\
\midrule
Range of area ratio to the largest & \((0,\frac{1}{4^8}]\) &  \((\frac{1}{4^8},\frac{1}{4^7}]\) & ... & \((\frac{1}{4^2},\frac{1}{4^1}]\) & \((\frac{1}{4^1},1]\) \\
Num of sampled points & \(1\) & \(4\) & ... & \(4^7\) & \(4^8\) \\
\bottomrule
\end{tabular}
}
\end{table}

Generally, the raw meshes do not offer information such as colors and textures, but we need to assign an initial color for the sampled points. It can be a completely random color for every point that achieves lower quality in the end, but there are some better methods. Thus, we apply a simple K-means \cite{macqueen1967some} clustering algorithm to the pixel colors of the image dataset, for picking up the \(k\)th (\(k = 3\)) most representative color of the dataset. It is done by calculating the cluster center, or the mean value of the colors in the \(k\)th cluster:
\begin{equation}
    \mathbf{S} = \argmin_{\mathbf{S}} \sum_{i=1}^{k} \sum_{\mathbf{c} \in \mathbf{S_i}} ||\mathbf{c}-\mathbf{\mu_i}||^2, 
\end{equation}
\begin{equation}
    \mathbf{\mu_i} = \frac{1}{|\mathbf{S_i}|} \sum_{\mathbf{c} \in \mathbf{S_i}} \mathbf{c}, 
\end{equation}
where \(\mathbf{S}\) is the cluster of colors and \(\mathbf{\mu_i}\) is the most representative color. We randomly choose one color from them and assign it to the initial color of the 3D Gaussians. This helps to represent the mesh surfaces more accurately.

\begin{figure}
    \centering
    \includegraphics[width=1\linewidth]{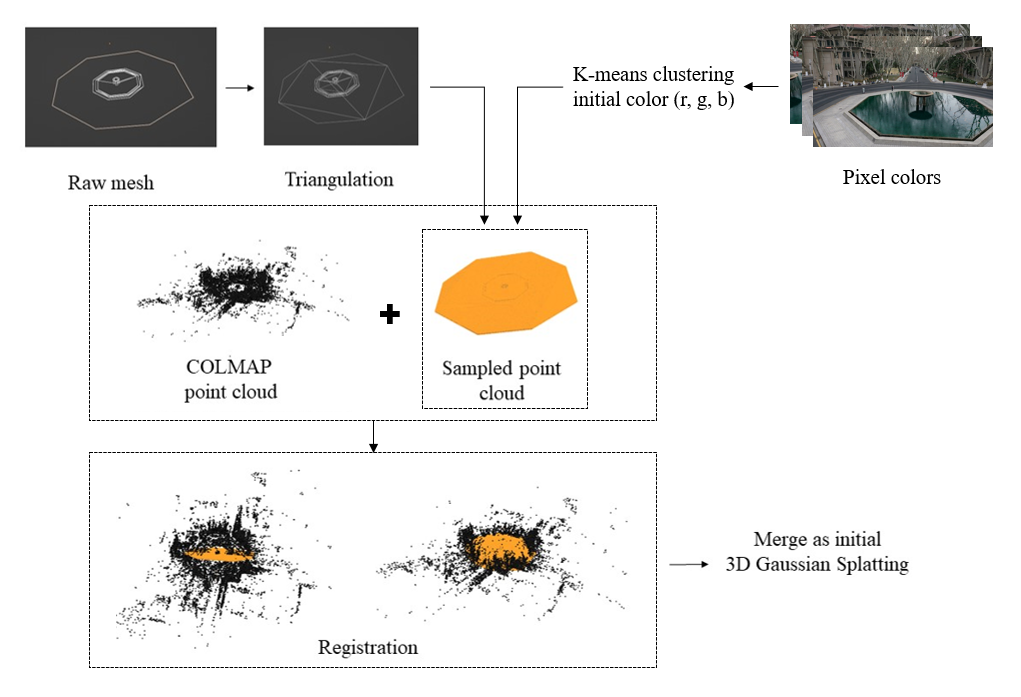}
    \caption{The pipeline of our method.}
    \label{fig:fig3}
\end{figure}

In this way, we have completed the sampling of evenly spaced points on the mesh surface and subsequent computations can be performed. The full pipeline of our straightforward approach of leveraging raw models to enhance 3D Gaussian splatting is shown in \cref{fig:fig3}.

\section{Experiment}

\subsection{Data preparation}
\begin{figure}
    \centering
    \includegraphics[width=1\linewidth]{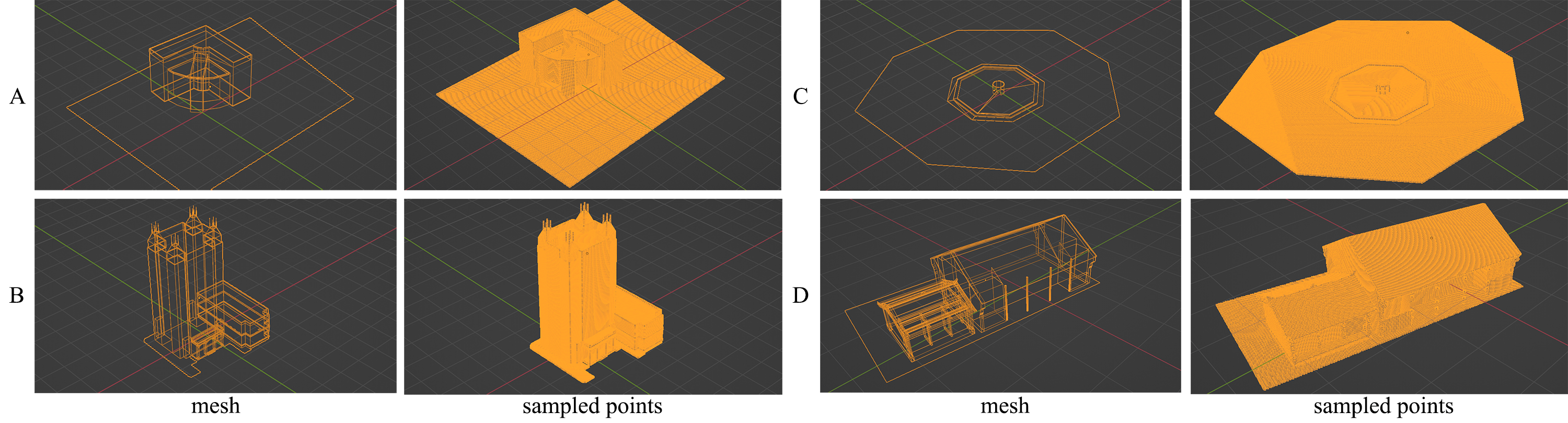}
    \caption{Demonstration of raw mesh models and point clouds sampled with our method of each scene in Blender.}
    \label{fig:fig4}
\end{figure}

\begin{figure}
    \centering
    \includegraphics[width=0.8\linewidth]{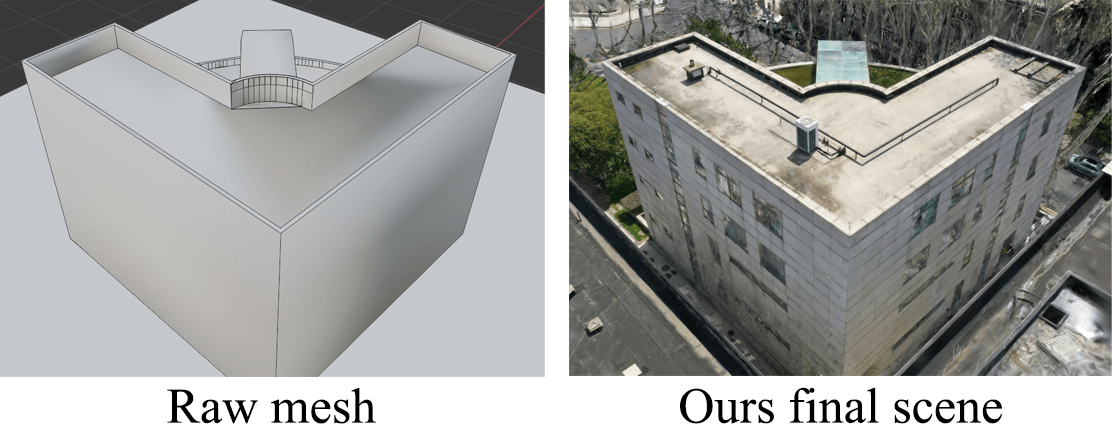}
    \caption{Comparison between the raw model with no detail and the scene rendered with our output. It is obvious that the mesh provides no detailed structures, and they must be reconstructed during training of the 3DGS.}
    \label{fig:fig5}
\end{figure}

We leverage a coarse approximation from a low-polygon mesh, which does not contain details. They can be constructed by measuring key points of the buildings. The raw mesh models we acquired are sourced from the database of the Architectural Research Institute of cooperative units. Before input, the number of mesh triangles is ensured to be within the range between 100 and 1000 (exact parameters in \cref{tab:tabrawmesh}), resulting in our sample points approximately ranging from 300k to 1M in quantity. The raw mesh models we utilize are illustrated in \cref{fig:fig4}. We provide a comparison between the example raw model and the final 3DGS scene as shown in \cref{fig:fig5} to show how simple the raw models are. Please note that those raw models only contain the general shape of the architectural structure and the details are significantly different from the final scene or actual objects. These detailed attributes must be completed by 3DGS training.

\begin{table}
    \caption{Element counts of raw mesh}
    \label{tab:tabrawmesh}
    \centering
    \begin{tabular}{c|c|c|c|c}
        \hline
        Count/Scene & A & B & C & D \\
        \hline
        Vertex & 294 & 1,965 & 759 & 1,135 \\
        Face & 152 & 491 & 170 & 780 \\
        \hline
    \end{tabular}
\end{table}

\begin{figure}
    \centering
    \includegraphics[width=1\linewidth]{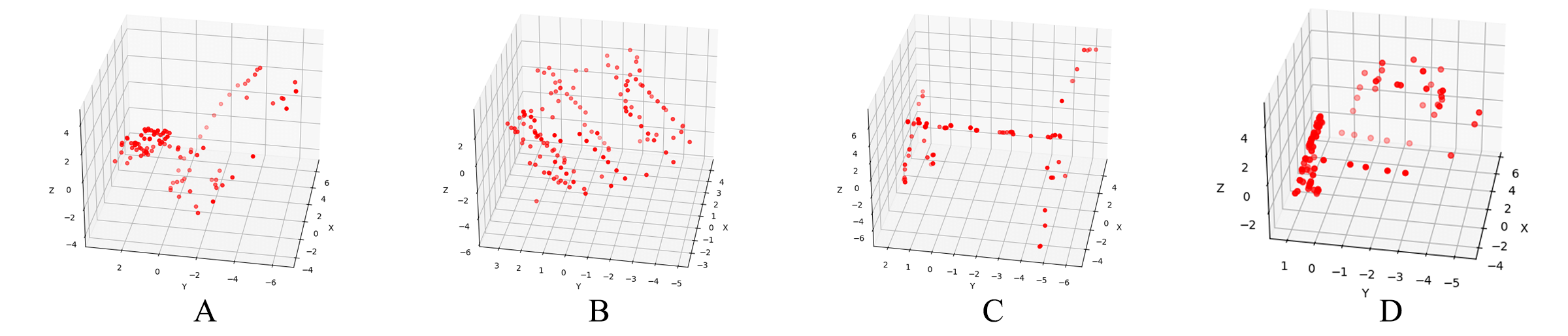}
    \caption{Camera viewpoints of scenes A, B, C, D.}
    \label{fig:fig6}
\end{figure}

We collect our sample dataset of the building photographs using DJI Mavic2. The dataset includes four architectural scenes - A, B, C, and D, and each architectural scene is comprised of 75 to 165 images. There is no fixed flight trajectory or maintained distance from the target building, in addition to that any of the scenes may contain areas obstructed by trees or other buildings that limit the flight route of the drone and block certain views from being captured by the camera, thus causing some photographs to be taken from inconvenient angles. The camera positions of the dataset are shown in \cref{fig:fig6}. Please note that the uniformity of these camera positions is far less than that of the Mip-Nerf-360 dataset \cite{Barron_2022_CVPR} used in the original 3D Gaussian Splatting paper by Kerbl \etal \cite{kerbl3Dgaussians}. In the photo set of scene A, C, and D, some surfaces of the building might be captured on several separate photos, having better reference, while others are captured during a fly-through path over the target, as the drone is not able to obtain photos with sufficient field of view in narrow spaces between objects. In scene B, the viewpoints are concentrated on several lines going up along the building.

\subsection{Implementation and Results}
The execution of sampling points on the mesh surface is high-speed, requiring less than one second when sampling 1M points. We provide a comparison of execution speed against Open3D Poisson disk sampling in \cref{tab:tabexecutiontime}. Despite their potential for similar quality, we opted not to introduce time-consuming computations at this stage to preserve the lightweight nature of our pipeline. The sampled points are shown in \cref{fig:fig4}. Our K-means clustering for pixel colors takes an average of 30 seconds for image datasets reshaped to 140p resolution (enough for picking up color).

\begin{table}
    \caption{Execution time of points sampling}
    \label{tab:tabexecutiontime}
    \centering
    \begin{tabular}{c|c|c|c|c}
        \hline
          & A & B & C & D \\
          \hline
        Number of Points & 339,622 & 388,895 & 329,783 & 414,244 \\
        Ours & 132ms & 125ms & 142ms & 178ms \\
       o3d Poisson disk sampling & 2 min 32.3 s & 3 min 48.3 s & 2 min 18.4 s & 4 min 18.3 s \\
        \hline
    \end{tabular}
\end{table}

All photos are resized to a size of 1920×1080 pixels for COLMAP and 3DGS training. To process the initial information of the raw model, we manually performed coarse registration for the sampled point cloud and the COLMAP point cloud in Blender, after which the point clouds could be merged. The criterion for performing point cloud registration is to perform rotation, translation, and XYZ axis scaling to align the ground and wall’s direction and position. Due to the powerful reconstruction capability of the 3DGS, the registration does not require fine-tuning by an algorithm such as ICP.

Our initial information does not significantly impact the running speed of the 3D Gaussians, meaning it can still complete 30,000 iterations of training in 20 to 30 minutes and achieve real-time rendering. For the parameters of the 3D Gaussian splatting model, we use the same settings as in the original paper, and all models are trained for 30,000 iterations. The experiment is conducted on a server with an Intel® Xeon® Silver 4216 CPU @ 2.10GHz and an NVIDIA RTX 3090 24GB GPU.

\begin{figure}
    \centering
    \includegraphics[width=1\linewidth]{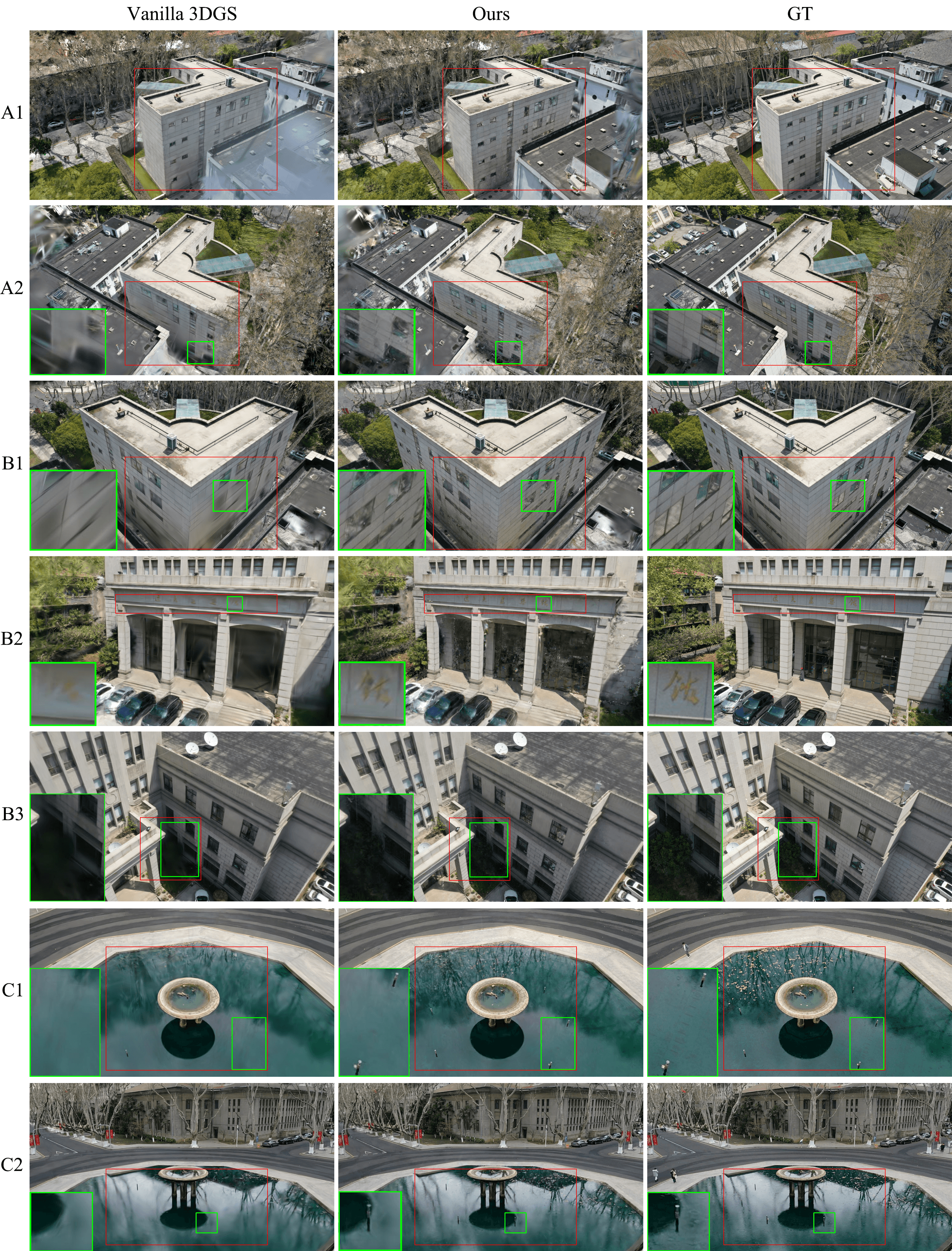}
    \caption{Qualitative comparison among vanilla 3D Gaussian Splatting, our method, and the Ground Truth.}
    \label{fig:fig7}
\end{figure}

\begin{figure}
    \centering
    \includegraphics[width=1\linewidth]{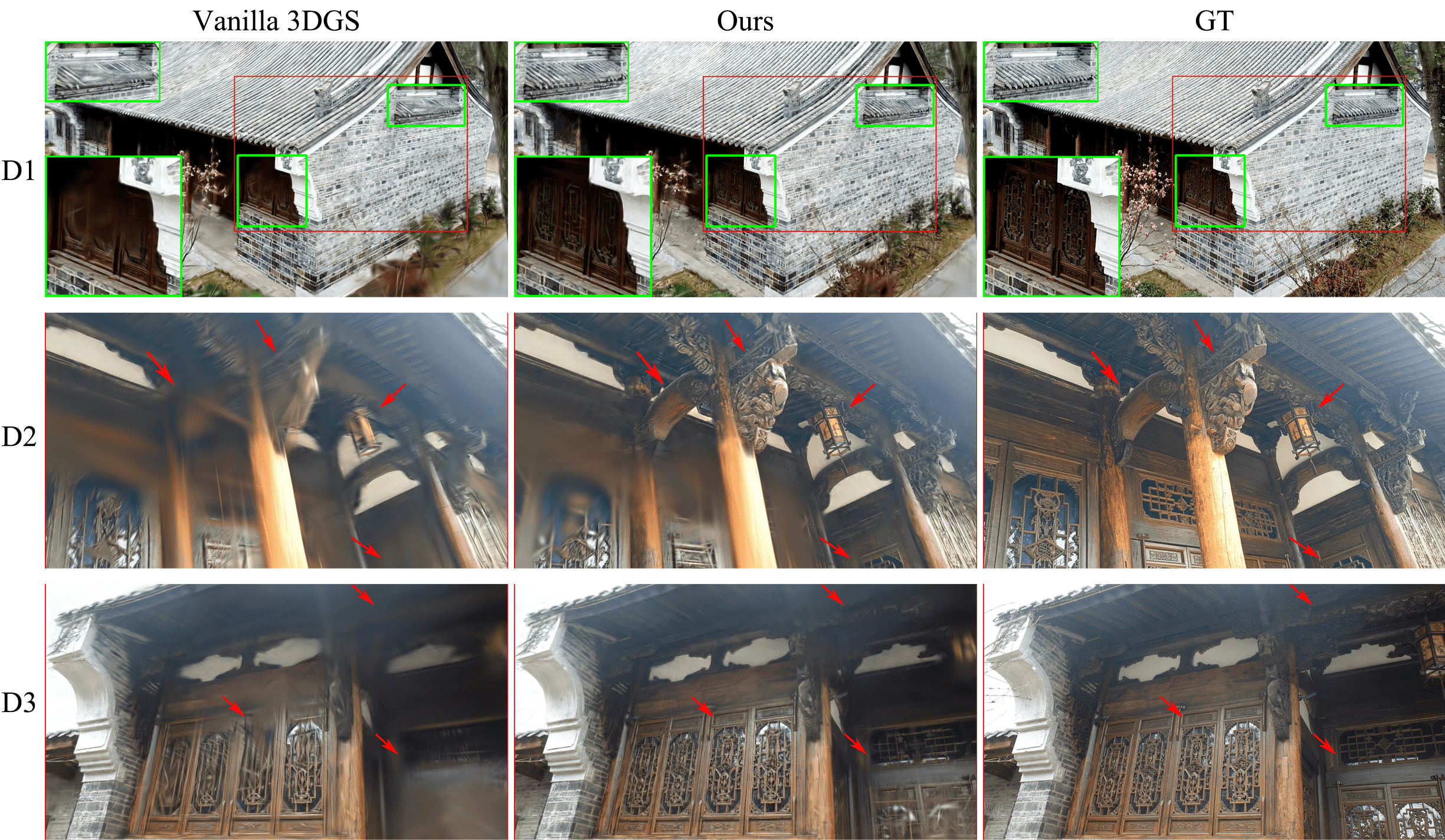}
    \caption{Figure 7 \(cont.\).}
    \label{fig:fig7cont}
\end{figure}

The experiment results of our validation set images are displayed in \cref{fig:fig7} and \cref{fig:fig7cont}. We only demonstrate the visual quality of the target building and the surroundings included in the mesh, because meshes are the key in our method. Thus the portion of the image within the red bounding box (full image in D2 and D3) is used for the computation of Structural Similarity Index (SSIM) and Peak Signal-to-Noise Ratio (PSNR) in \cref{tab:Table2}. Please take note of the significant difference between our results and the vanilla method in the green box.

\begin{table}[h]
\caption{PSNR and SSIM metrics compared between two methods}
\label{tab:Table2}
\centering
\begin{tabular}{ m{6em}<{\centering} | m{2cm}<{\centering} | m{2cm}<{\centering} | m{2cm}<{\centering} | m{2cm}<{\centering} } 
\hline
\multirow{2}{6em}{Scene} & \multicolumn{2}{c|}{SSIM$\uparrow$} & \multicolumn{2}{c}{PSNR$\uparrow$} \\
\cline{2-5}
& Ours & vanilla 3DGS & Ours & vanilla 3DGS \\
\hline
A & 0.7604 & 0.7314 & 23.57 & 22.21 \\
B & 0.8481 & 0.8345 & 26.63 & 25.75 \\
C & 0.8563 & 0.8030 & 26.67 & 24.88 \\
D & 0.7193 & 0.6950 & 24.25 & 23.06 \\
\hline
\end{tabular}
\end{table}

In result A, the vanilla 3D Gaussian Splatting method, failed to capture the wall details, rendering the wall itself but losing the windows on it due to certain images being captured from angles with a poor view and less reference. After enhancing the original code with the raw model, we have successfully rendered the object in the correct shape. In result B, vanilla 3D Gaussian failed to capture either large characters above the gate or the details of the doors and windows at the entrance of the building. Large and over-blurry Gaussians are generated in vanilla 3DGS, while after our enhancement, smaller Gaussians make the characters and tiny structures clearer. In result C, the fountain pipes are not visible from various angles using vanilla 3D Gaussian Splatting. This suggests that the method fails to capture this detail across all views. However, upon importing the raw model, the pipes become visible in all directions, indicating improved shape and texture-capturing capabilities after enhancement. In result D, an ancient-style building is depicted. However, vanilla 3D Gaussian Splatting falls short of capturing the texture on the windows, pillars, lanterns, and tiles on the eaves. Fortunately, after enhancement, these details become clearly visible.

\section{Discussion and Further Work}
The point cloud generated by SfM determines the initial positions of Gaussians. This process subsequently undergoes a densification or pruning of these Gaussian points, from iteration 500 to iteration 15,000. The 3D Gaussian Splatting algorithm learns the geometry of the scene by gradually densifying those Gaussians throughout the entire scene during training. When a Gaussian point is present within the scene, it learns other attributes, such as position, opacity, and color. However, if the image dataset fails to cover a specific area adequately, the initial point cloud calculated by SfM will also exhibit subpar coverage of that area. Also, because the method trains on full images, less coverage leads to less contribution to loss and gradients in this area during training, posing challenges for 3D Gaussians to yield high-quality results.

We directly import the initial point cloud sampled from the raw model. This ensures that the Gaussians, each representing a more detailed part of the scene, can achieve satisfactory coverage even without densification in this area. Furthermore, these initially sampled point clouds participate in a significantly larger number of iterations during the training process, thereby capturing textures of the details with superior quality.

\begin{figure}
    \centering
    \includegraphics[width=1\linewidth]{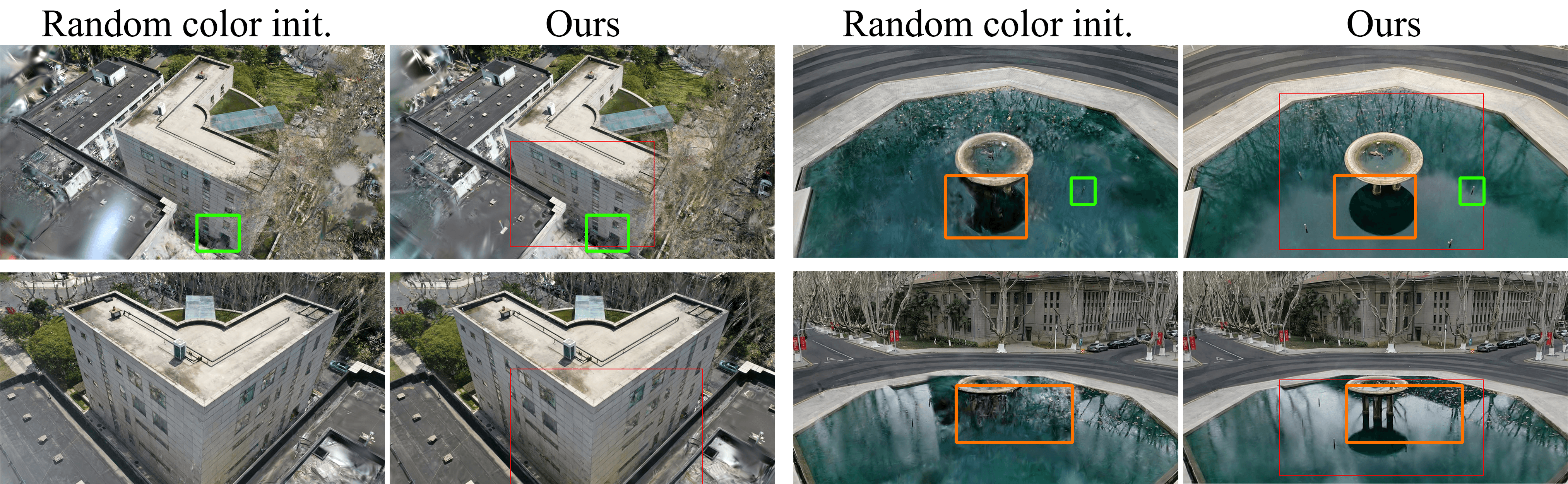}
    \caption{Random initialization vs. ours}
    \label{fig:figreb}
\end{figure}

We compared our method with random color initialization in \cref{fig:figreb}. While random initialization produces similar results in some scenes and regions (\eg, green box), it introduces more noise in areas with sparse views, significantly reducing the effect (\eg., orange box). This highlights the importance of our color initialization. 

Our manual registration of point clouds may introduce some errors, and its accuracy is unlikely to match automated registration, particularly when the acquisition device is calibrated. Therefore, we consider the performance reported to be a lower bound of the method, with potential for improvement through refined registration. Thus, we believe that despite some parts of the pipeline being performed manually, it does not undermine the demonstration that the overall process (i.e., the 3DGS can be enhanced by introducing mesh sampled point clouds) is effective.

Additionally, our data were gathered during a practical engineering process, resulting in varying degrees of complexity and precision. Nevertheless, our results demonstrated that our enhancement has led to notable improvements in performance across various scenarios, suggesting that our method possesses sufficient stability.

Our approach, while effective, is not without limitations. One such challenge is the determination of the optimal quantity of points sampled on the raw model surface, \ie, the number of initial 3D Gaussians. The selection is currently performed manually; however, the process could be enhanced by applying adaptive methods or filters \cite{jung2024relaxing}. In certain regions, an excessively large number of initial points can lead to a surface that is rendered with noise, while on the other hand, an overly small number of sampled points may not yield a significant improvement. 

\begin{figure}
    \centering
    \includegraphics[width=0.9\linewidth]{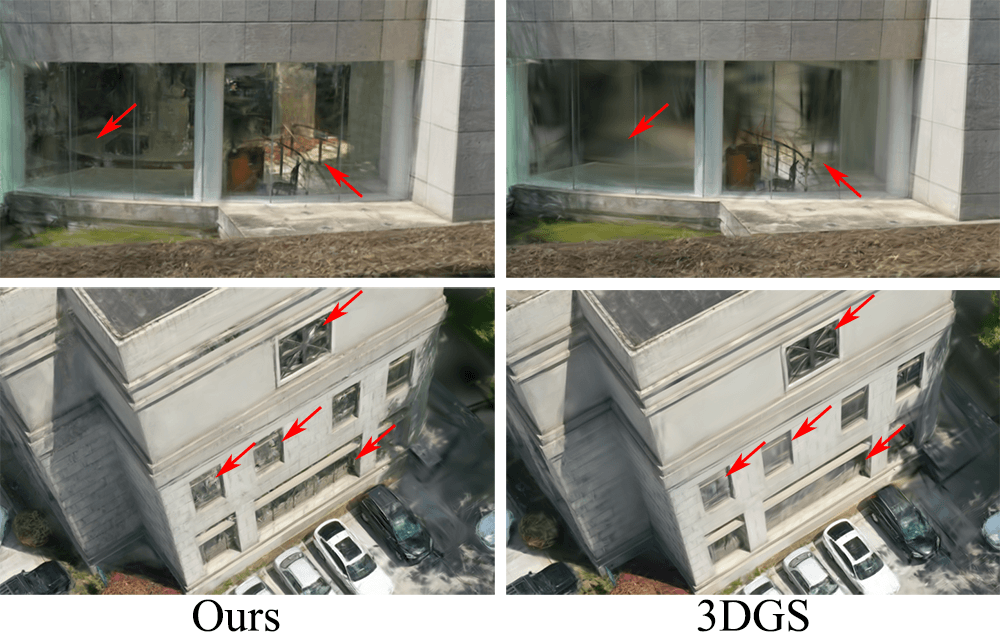}
    \caption{Rendered results of reconstruction behind transparent and specular materials. Our method can improve the visual quality if no specular highlights are captured (small incident angle, first row), but still struggles with reflections (large incident angle, second row). 3DGS without accurate initial points tends to produce over-blurry Gaussians in both cases.}
    \label{fig:fig8}
\end{figure}

Further investigation is needed to explore a better method for choosing areas of point sampling in regions obscured by transparent and specular materials. Those areas with transparent materials behave differently when viewed from various directions in the image dataset, thus highlights cause problems if being rendered (Note that rendering is a step of reconstructing from the images.) at a large incident angle (\cref{fig:fig8}). For instance, in the case of Building A, the inclusion of initial points on the ground behind the window results in improved texture quality. However, for Building B, points sampled on the ground behind the windows within the building can mislead the method to mix the foreground and background in the windows during rendering, leading to problems learning the correct structure. In contrast, the vanilla 3DGS tends to produce over-blurry Gaussians in those specular areas. It can be addressed by a more nuanced method for choosing areas of point sampling or adopting an advanced technique in 3DGS training for reflections\cite{jiang2023gaussianshader3dgaussiansplatting, yang2024specgaussiananisotropicviewdependentappearance, comi2024snapittapitsplatittactileinformed}.

\section{Conclusion}
This paper introduces a straightforward strategy to utilize raw models for guiding 3D Gaussian Splatting, thereby enhancing its visual quality when applied in the architectural domain. Our findings indicate that when 3D Gaussian Splatting is implemented in a complex building scene, the shape and texture of surfaces remain of low quality if the photos have limited coverage. Despite 3D Gaussian Splatting being a state-of-the-art technique for rendering photo-realistic scenes of objects, it does not exploit the geometric prior knowledge inherent to specific fields.

To address this, we sample an initial point cloud on the surface of raw models of the corresponding building, assign the initial color by clustering pixel colors, and align it with the point cloud output by COLMAP. This assists 3D Gaussian Splatting in capturing the texture and more detailed structures. We implemented this approach in four scenes with photos captured in a non-
systematic manner, significantly improving geometry and texture quality. This underscores the potential of our proposed method in applying to the architectural industry.

\bibliographystyle{splncs04}
\bibliography{refs}

\end{document}